\documentclass[sigconf]{acmart}

\AtBeginDocument{%
  \providecommand\BibTeX{{%
    \normalfont B\kern-0.5em{\scshape i\kern-0.25em b}\kern-0.8em\TeX}}}

\setcopyright{rightsretained}
\acmConference[COLIEE 2021]{COLIEE 2021 workshop: Competition on Legal
  Information Extraction/Entailment }{June 21, 2021}{Online}
\acmISBN{}
\acmDOI{}

\renewcommand\footnotetextcopyrightpermission[1]{}

\usepackage[normalem]{ulem}
\usepackage{CJKutf8} 
\useunder{\uline}{\ul}{}

\begin{document}

\title{JNLP Team: Deep Learning Approaches for Legal Processing Tasks in COLIEE 2021}


\author{Ha-Thanh Nguyen}
\affiliation{%
  \institution{Japan Advanced Institute of Science and Technology}
  \city{Ishikawa}
  \country{Japan}
}

\author{Phuong Minh Nguyen}
\affiliation{%
  \institution{Japan Advanced Institute of Science and Technology}
  \city{Ishikawa}
  \country{Japan}
}

\author{Thi-Hai-Yen Vuong}
\affiliation{%
  \institution{University of Engineering and Technology, VNU}
  \city{Hanoi}
  \country{Vietnam}
}

\author{Quan Minh Bui} 
\affiliation{%
  \institution{Japan Advanced Institute of Science and Technology}
  \city{Ishikawa}
  \country{Japan}
}

\author{Chau Minh Nguyen} 
\affiliation{%
  \institution{Japan Advanced Institute of Science and Technology}
  \city{Ishikawa}
  \country{Japan}
}

\author{Binh Tran Dang}
\affiliation{%
  \institution{Japan Advanced Institute of Science and Technology}
  \city{Ishikawa}
  \country{Japan}
}

\author{Vu Tran} 
\affiliation{%
  \institution{Japan Advanced Institute of Science and Technology}
  \city{Ishikawa}
  \country{Japan}
}

\author{Minh Le Nguyen}
\affiliation{%
  \institution{Japan Advanced Institute of Science and Technology}
  \city{Ishikawa}
  \country{Japan}
}


\author{Ken Satoh}
\affiliation{%
  \institution{National Institute of Informatics}
  \city{Tokyo}
  \country{Japan}
}

\renewcommand{\shortauthors}{JNLP Team}

\begin{abstract}
  COLIEE is an annual competition in automatic computerized legal text processing.
  Automatic legal document processing is an ambitious goal, and the structure and semantics of the law are often far more complex than everyday language.
  In this article, we survey and report our methods and experimental results in using deep learning in legal document processing.
  The results show the difficulties as well as potentials in this family of approaches.
\end{abstract}

\ccsdesc[500]{Computing methodologies~Neural networks}
\ccsdesc[300]{Applied computing~Law}

\keywords{Deep Learning, Legal Text Processing, JNLP Team}


\maketitle

\section{Introduction}
COLIEE is an annual competition in automatic legal text processing.
The competition uses two main types of data: case law and statute law.
The tasks for automated models include: retrieval, entailment, and question answering.
With deep learning models, JNLP team achieves competitive results in COLIEE-2021.

Task 1 is a case law retrieval problem. With a given case law, the model needs to extract the cases that support it. This is an important problem in practice. It is actually used in the attorney's litigation as well as the court's decision-making.
Task 2 also uses caselaw data, though, the models need to find the paragraphs in the existing cases that entail the decision of a given case.
Task 3, 4, 5 uses statute law data with challenges of retrieval, entailment, and question answering, respectively.

For the traditional deep learning approach, the amount of data provided by the organizer is difficult for constructing effective models. For that reason, we use pretrained models from problems that have much more data and then finetune them for the current task.
In Tasks 1, 2, 3, and 4 we have used lexical score and semantic score to filter a correct candidate. In our experiments, the ratio is not 50:50 for lexical score and semantic score, we find out that the increase in the rate of lexical score lead to efficiency in ranking candidates.

Through problem analysis, we propose solutions using deep learning. We also conducted detailed experiments to explore and evaluate our approaches.
Our proposals of deep learning methods for these tasks can be a useful reference for researchers and engineers in automated legal document processing.

\begin{figure*}
    \centering
    \includegraphics[scale=0.5]{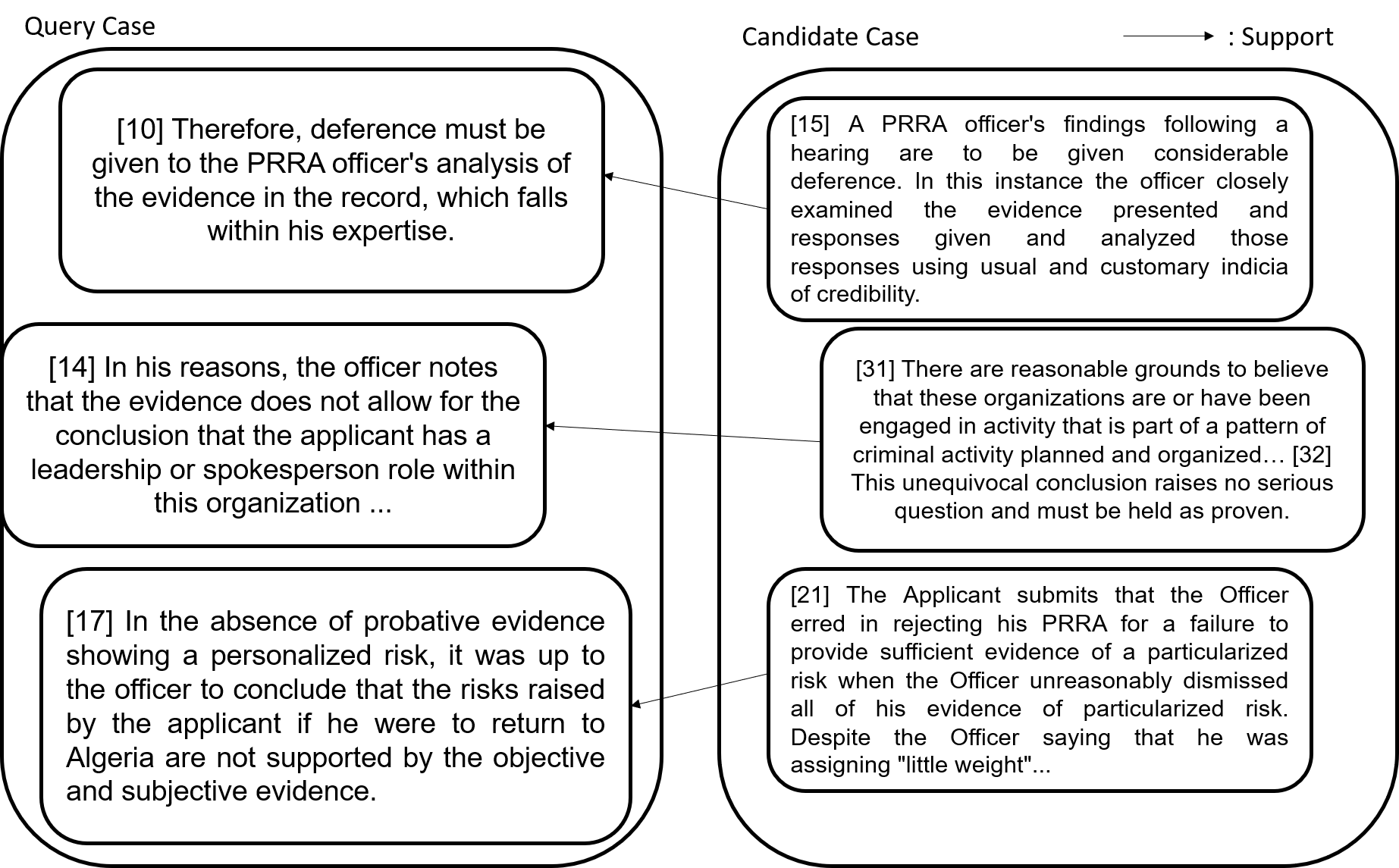}
    \caption{Supporting Definition.}
    \label{supporting_definition}
\end{figure*}

\section{Related Works}
\subsection{Case Law} 
In COLIEE 2018, most teams chose the lexical-based approaches. UBIRLED ranked the candidate cases based on tf-idf. They got approximately 25\% of the candidate cases with the highest scores. UA and several teams also chose the same approach with UBIRLED. They compared lexical features between a given base case and corresponding candidate cases. JNLP team combined lexical matching and deep learning, which achieved state-of-the-art performance on Task 1 with the F1 score of 0.6545.

In COLIEE 2019, several teams applied machine learning including deep learning to both tasks. JNLP team achieved the best result of Task 1 in COLIEE 2019 \cite{tran2019building} using an approach similar to theirs in COLIEE 2018. In Task 2, their deep learning approach achieved lower performance compared to their lexical approach. Team UA's combination of lexical similarity and BERT model achieved the best performance for Task 2 in COLIEE 2019~\cite{10.1145/3322640.3326741}.

In COLIEE 2020, the Transformer model and its modified versions were widely used. TLIR and JNLP~\cite{nguyen2020jnlp} teams used them to classify candidate cases with two labels (support/non-support) in Task 1. Team cyber encoded candidate cases and the base case in tf-idf space and used SVM to classify. They demonstrated the ability of the approach with the first rank in Task~1. In Task 2, JNLP~\cite{nguyen2020jnlp} continually applied the same approach in Task 1 in the weakly-labeled dataset. It made them surpass team cyber and won Task~2.
 
\subsection{Statute Law}
The retrieval task (Task 3) is often considered as a ranking problem with similarity features.
In COLIEE 2019, most of the teams used lexical methods for calculating the relevant scores. JNLP~\cite{jnlp_3_2019}, DBSE~\cite{dbse_3_2019} and IITP~\cite{iitp_3_2019} chose tf-idf and BM25 to build their models. JNLP used tf-idf of noun phrase and verb phrase as keywords which show the meaning of statements in the cosine-similarity equation. DBSE applied BM25 and Word2vec to encode statements and articles. The document embedding was used to calculate and rank the similarity score. Team KIS~\cite{kis_3_2019} represented the article and query as a vector by generating a document embedding. Keywords were selected by tf-idf and assigned with high weights in the embedding process. In COLIEE 2020, with the popularity of Transformer based methods, the participants change their approaches. The task winner, LLNTU, only used BERT model to classify articles as relevant or not.
    
Regarding the entailment task (Task 4), approaches using deep learning have attracted more attention. In COLIEE 2019, KIS~\cite{kis_4_2019} used predicate-argument structure to evaluate similarity. IITP~\cite{iitp_3_2019} and TR~\cite{tr_4_2019} applied BERT for this task. JNLP~\cite{jnlp_task4_coliee2019} classified each query to follow binary classification based on big data. In COLIEE 2020, BERT and multiple modified versions of BERT were used. JNLP~\cite{nguyen2020jnlp} chose a pretrained BERT model on a large legal corpus to predict the correctness of statements.

\begin{figure*}
    \centering
    \includegraphics[scale=0.5]{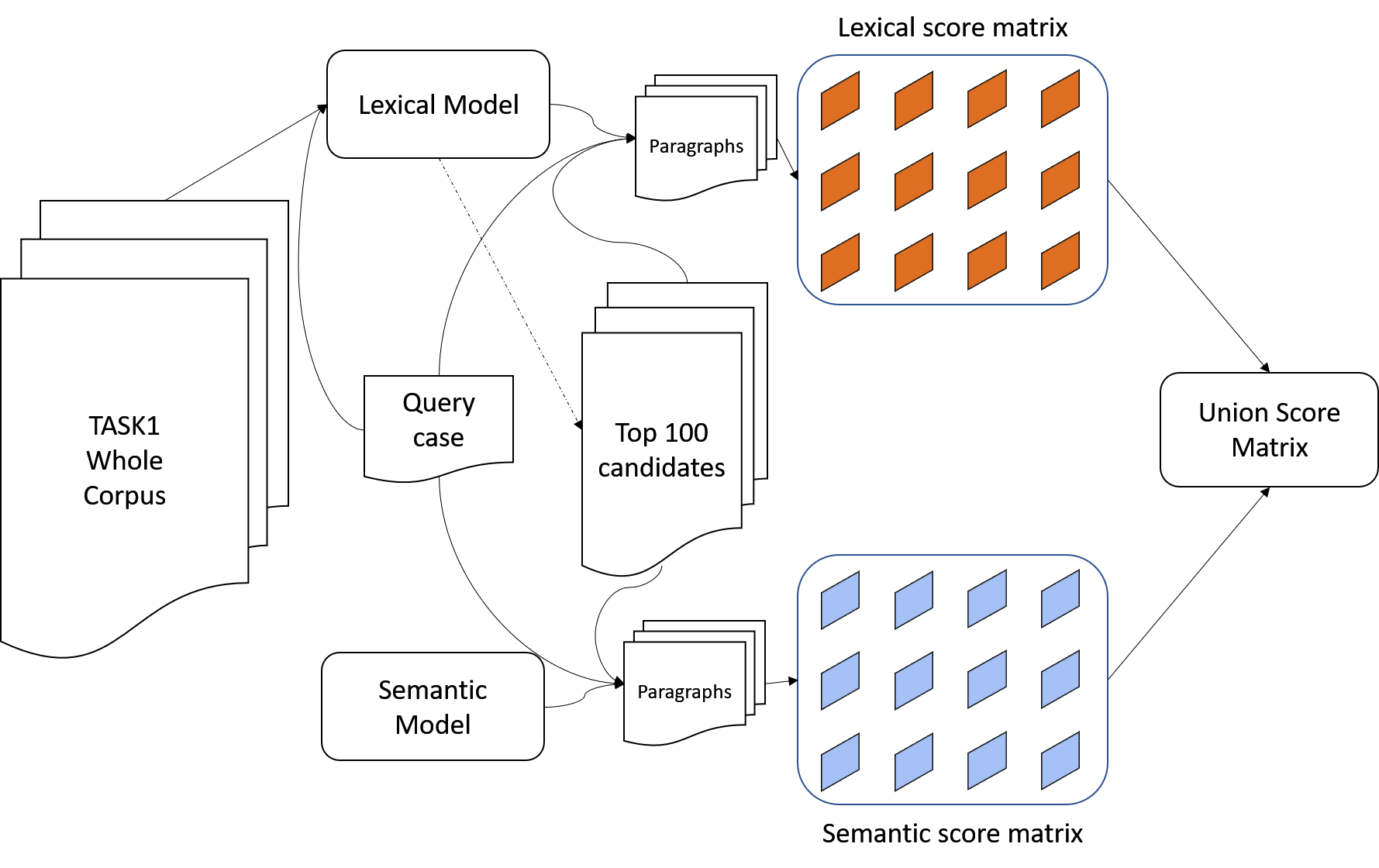}
    \caption{Demonstration of mixing lexical and semantic score.}
    \label{Union_score}
\end{figure*}

\section{Method}
\label{sec:method}

\subsection{Task 1 and Task 2. Case Law Processing}
In 2021, COLIEE has changed a lot in the data structure of Task 1, and this modification made this task more challenging. In the last year, for each given query, there are only about 200 candidate cases to search for the relevant case law. But in this year, for a given query, we have to search over 4000 candidate cases for relevant cases. Because of the competition's increment of difficulty, the performance of participants' models drops badly from nearly 70\% to 20\% or worse. The particular reason for this circumstance is the huge searching space. For tackling this issue, we firstly used a lexical model to calculate the similarity between a given query and the whole case law corpus. Limiting the searching space from 4000 candidate cases to 100 candidate cases by picking the top 100 cases that have the highest BM25 scores for a given query.

The previous approaches focus on encoding the whole case law and calculating the similarity of vector representation for each query-candidate pair of cases. In our method, we handle the similarity between candidates on the paragraph level. "Supporting" and "Relevant" are subtractive definitions, and we assume that for each paragraph in the query case, there are one or more paragraphs that carry useful information for the given query's paragraph. As we can see in the Figure \ref{supporting_definition}, paragraph [10] has \textit{"PRRA officer's"} and in the candidate case number [15] also has \textit{"PRRA officer's"}, there seems to be a lexical relationship here. This is evidence that the lexical model could be effective. Along the line of the query paragraph number [14] has \textit{"the evidence"} corresponding to \textit{reasonable grounds} in the candidate paragraph number [31], we can see the semantic relationship between two paragraphs through this example. For discovering the relevant paragraph, we combine the lexical similarity and the semantic similarity score.

\textbf{Lexical Matching} For the exact purpose of capturing lexical information, we use Rank-BM25~\footnote{https://pypi.org/project/rank-bm25/}, a collection of algorithms for querying a set of documents and returning the ones most relevant to the query. For Task 1, after reducing the searching space to 100 queries, we separate the query case and also candidate cases into paragraphs. Assume that the query case has $N$ paragraphs, and the candidate case has $M$ paragraphs, we will calculate the lexical mapping score by Rank-BM25 for each query paragraph and every single paragraph in the candidate case. Then we obtain the matrix lexical score size $N x M$ for each query-candidate pair. We keep these matrixes and their union with supporting score which is introduced later. 

\textbf{Supporting Matching} As we mentioned in the Figure \ref{supporting_definition}, we want to extract the semantic relationship between query paragraph and candidate paragraph. To obtain this relationship score, we use pretrained model BERT, and this model is provided by huggingface\footnote{https://huggingface.co/models}. The same approach as lexical matching, we sequentially split a query and corresponding candidate cases into paragraphs, and obtain a matrix semantic score that has the same size as matrix lexical score. For more details about this supporting model, we do not use the original model provided by huggingface. Although BERT is currently one of the best, it is trained in the general domain. This actively demonstrates that BERT can not work well in a specialized domain such as law. For tackling this issue, we develop a silver dataset based on the COLIEE dataset for finetuning BERT and use this for predicting the semantic relationship between paragraphs.

\textbf{Union Score} To obtain the most relevant cases for a given query, we find a semantic as well as lexical relationship overlapped between the two paragraphs. For the exact purpose of developing union score we use the following formula:
\begin{equation} \label{combine_score}
    union\_score = \alpha * score_{supporting} + (1-\alpha) * score_{BM25}
\end{equation}
Figure \ref{Union_score} shows the architecture of the system.

For Task 2, we use the same approach as Task 1. For the final runs, we use the supporting model and lexical model for 2 runs, and in the last run, we use NFSP model~\cite{jnlp_task5_coliee2021}. Using our proposed approach, Task2 is basically a binary classification with the training data as a set of sentence pairs. As a result, we can obtain more gold training data to train the supporting model. For optimizing the performance of the supporting model, after finetuning on silver, we finetune one more time on gold data of Task 2.

\subsection{Task 3. The Statute Law Retrieval Task\label{subs:3}} This task involves reading a legal bar exam question $Q$, and extracting a subset of Japanese Civil Code Articles $A_{i  \mid 1 \leq i \leq n} $ which contains appropriate articles for answering the question, i.e. Entails($A_{i  \mid 1 \leq i \leq n}$, $Q$) or Entails($A_{i  \mid 1 \leq i \leq n}$, \textit{not} $Q$).

As specified in \cite{nguyen2020jnlp}, two main challenges of Task 3 are: (i) answering the questions which describe specific legal cases (note that the language used in statute law tends to be general), and (ii) addressing the long articles. The first challenge requires the model to have deduction ability, which is a difficult task and requires further research. Regarding the second challenge, we addressed it by performing text chunking technique on the prepared training data and self-labeled technique while finetuning pretrained models.

\paragraph{\textbf{Training data preparation.}}
We follow the training data preparation method proposed in \cite{nguyen2020jnlp}. Naturally, the pair of a question and each of its annotated entailing articles is considered a positive training example, and the pair of a question and any other article is considered a negative training example. However, this approach results in very big negative:positive ratio in training data. To reduce this ratio, in this phase, the maximum number of negative training examples is limited to be 150, choosing based on the rank of tf-idf scores. Please refer to \cite{nguyen2020jnlp} for more details.

\paragraph{\textbf{Text chunking technique.}}
We use BERT and RoBERTa, which are two prevalent language models, as pretrained models. However, they cannot handle a very long article without truncating it, because of the 512-token limitation. Besides, in Task 3, we observe that, in most cases, only a few parts of the entailing article entails the corresponding question while other parts do not. We provide an example in Table~\ref{tab:example_chunk}, where the question is entailed by only one part of the article. To address the aforementioned limitation of BERT and RoBERTa, we propose to split each article into multiple chunks using a sliding window (as sliding windows mitigate the cases where the entailing part of the article is split). Regarding training data generation, we followed the method proposed in \citet{nguyen2020jnlp}, except that the pair used for training is ($question$, $chunk$) instead of ($question$, $article$). The label of pair ($question$, $chunk$) is derived from the label of its corresponding pair ($question$, $article$).

\begin{table}[!htbp]

\begin{tabular}{lp{0.35\textwidth}}

\toprule
\textbf{Q. R01-4-E}          & In cases any party who will suffer any detriment as a result of the fulfillment of a condition intentionally prevents the fulfillment of such condition, the counterparty may deem that such condition has been fulfilled. \\\midrule
\textbf{Article 130}          &  Part I General Provisions Chapter V Juridical Acts Section 5 Conditions and Time Limits  \\ & (Prevention of Fulfillment of Conditions)          \\ & \textbf{(1) If a party that would suffer a detriment as a result of the fulfillment of a condition intentionally prevents the fulfillment of that condition, the counterparty may deem that the condition has been fulfilled.}
\\ & (2) If a party who would enjoy a benefit as a result of the fulfillment of a condition wrongfully has that condition fulfilled, the counterparty may deem that the condition has not been fulfilled. \\
\bottomrule
\end{tabular}

\caption{An example that only one part of the entailing article entails the corresponding question. \label{tab:example_chunk}}
\end{table}
 
\paragraph{\textbf{Self-labeled technique.}}
Training data generated using text chunking contains noises. For example, assume that a long article $A$, which entails a question $Q$, is split into sub-articles $A_{1}, A_{2}, …, A_{n}$, and only $A_{n}$ entails $Q$, then the aforementioned training data generating method will label pairs ($Q, A_{1}$), ($Q, A_{2}$), …, ($Q, A_{n}$) as positive training examples; however, only pair ($Q, A_{n}$) should be labeled positive, while other pairs, i.e. ($Q, A_{1}$), ($Q, A_{2}$), …, ($Q, A_{n-1}$), should be labeled as negative examples. Inspired by the self-labeled techniques \cite{triguero2015self}, we propose to deploy a simple self-labeled technique to help mitigate noisy training examples. Specifically, first, a pretrained model finetunes on the generated training data. Next, the finetuned model predicts labels for training examples. After that, we modify the labels based on rules, and the label-modified training data is used for the next finetuning phase. This self-labeled and finetuning process can be iterated multiple times. Regarding the label modifying rules, we only keep label modification if the label is converted from positive to negative. After label modification, the number of noisy examples tends to decrease, which allows the previous-phase finetuned model to learn from a more accurate data.

\paragraph{\textbf{Model ensembling.}}
The outcome of each model is the prediction based on the particular characteristics the models learned during training. Since each model has its advantages and disadvantages, model ensembling, an effective machine learning approach for incorporating models, appears to help produce better predictions. In our work, we ensemble models by deploying weighted aggregation on models' predictions. Prior to the ensemble process, the outputs of multiple models were scaled by min-max normalization method, so that they were standardized in the range [$0, 1$]. It ensures that the standardized outputs of each model have a similar impact on the final prediction results. We divided the dataset into training, development, and test set. The ensemble method weights were constructed from the development set and applied to the test set.


\subsection{Task 4. The Legal Textual Entailment Task}

This task involves the identification of an entailment relationship between relevant articles $A_{i  \mid 1 \leq i \leq n} $ (which is derived from Task 3's results) and a question $Q$. The models are required to determine whether the relevant articles entail "$Q$" or "$not Q$".  Given a pair of legal bar exam question and article $(Q, A_i)$, the models return a binary value for determining whether $(A_i)$ entails $(Q)$. To address this task, we modified the training data preparation step, then use thed same model architecture in Task 3 (Section~\ref{subs:3}) for training. 

\paragraph{Data preparation.} Based on our observation, the challenge of this task is to extract the relevance between a question and articles for classification while the number of given articles is relatively small (usually 1 or 2 articles are given). We hypothesize that the model can extract information more effectively and consistently if there are more relevant articles given. Therefore, we adapt the data augmentation technique mentioned in Task 3 (which is based on tf-idf scores) to increase the number of relevant articles for each question. Besides, we also use the \textit{text chunking} and \textit{self-labeled}   techniques introduced in Section~\ref{subs:3} for dealing with long article challenge.

\subsection{Task 5. Statute Law Question Answering}
The goal in Task 5 of the models is to answer legal questions. In detail, with a given statement, the model needs to answer whether the statement is true or false in the legal aspect.
In essence, Task 5 is constructed from Task 4, ignoring the step of retrieving the related clauses from the Japanese Civil Code.

Our novelty in Task 5's solution is the introduction of two NMSP and NFSP models.
The main idea in building these two models is to use translation information as means of ambiguity reduction.
We argue that a sentence in natural language can have many meanings, but in its translation, the most correct meaning will be expressed.
In addition, the meaning is also determined by the context, that is, the sentences before and after the current sentence.

These models, named ParaLaw Nets~\cite{jnlp_task5_coliee2021}, are pretrained on cross-lingual sentence-level tasks before being finetuned for use in the COLIEE problem.
The data we use to pretrain these models is bilingual Japanese law data provided by Japanese Law Translation website~\footnote{https://www.japaneselawtranslation.go.jp}.
We formulate the pretraining task for NFSP as a binary classification problem and NMSP as a multi-label classification problem.

We design the pretraining task to force the model to learn the semantic relationship in 2 continuous sentences crossing two languages.
From original sentences as "The weather is nice. Shall we go out?", their translations \begin{CJK*}{UTF8}{min}"いい天気ね。お出掛けしよ？"\end{CJK*}, and random sentences as "Random sentence.", \begin{CJK*}{UTF8}{min}"ランダム文。"\end{CJK*}, we can generate the training samples as in Table~\ref{tab:pretraining_samples}.

\begin{table}[]
\small
\begin{tabular}{lcc}
\hline
\textbf{Sentence Pair} & \textbf{NFSP Label} & \textbf{NMSP Label} \\
\hline
Shall we go out? The weather is nice.       & -                 & 2  \\
\begin{CJK*}{UTF8}{min}お出掛けしよ？いい天気ね。\end{CJK*}      & -                 & 2  \\
\begin{CJK*}{UTF8}{min}お出掛けしよ？\end{CJK*} The weather is nice.       & -                 & 2  \\
Shall we go out? \begin{CJK*}{UTF8}{min}いい天気ね。\end{CJK*}      & -                 & 2  \\
\hline
\begin{CJK*}{UTF8}{min}いい天気ね。お出掛けしよ？\end{CJK*}      & -                 & 1  \\
The weather is nice. Shall we go out?      & -                 & 1  \\
The weather is nice. \begin{CJK*}{UTF8}{min}お出掛けしよ？\end{CJK*}      & 1                 & 1  \\
\begin{CJK*}{UTF8}{min}いい天気ね。\end{CJK*}Shall we go out?      & 1                 & 1  \\
\hline
The weather is nice. \begin{CJK*}{UTF8}{min}ランダム文。\end{CJK*}      & 0                 & 0  \\
\begin{CJK*}{UTF8}{min}いい天気ね。\end{CJK*}Random Sentence.      & 0                 & 0  \\
The weather is nice. Random Sentence.      & -                 & 0 \\
\begin{CJK*}{UTF8}{min}いい天気ね。ランダム文。\end{CJK*}      & -                 & 0  \\
\hline
\end{tabular}
\caption{Examples of pretraining data.}
\label{tab:pretraining_samples}
\end{table}

These models are pretrained until the performance on the validation set does not increase.
Through the results in Table \ref{tab:pretraining_paramters}, we see that the models have better performance on the validation set with NFSP task, base models outperform distilled models. With these numbers, we can accept the assumption that the NFSP task is more straightforward than the NMSP task.

\begin{table}[]
\small
\begin{tabular}{lcccc}
\hline
\textbf{Model} & \textbf{Max Len.} & \textbf{Batch Size} & \textbf{\#Batches} & \textbf{Acc.} \\
\hline
NFSP Base       & 512                 & 16                  & 24,000                     & 94.4\%                       \\
NFSP Distilled  & 512                 & 32                  & 34,000                     & 92.2\%                       \\
NMSP Base       & 512                 & 16                  & 320,000                    & 88.0\%                       \\
NMSP Distilled  & 512                 & 32                  & 496,000                    & 87.7\%                       \\
\hline
\end{tabular}
\caption{Parameters and performances in pretraining the models on valid set.}
\label{tab:pretraining_paramters}
\end{table}

\begin{table}[]
\small
\begin{tabular}{ccccc}

\toprule
\textbf{Data Source}    & \textbf{\#case} & \textbf{\#paragraph} & \textbf{\#sentence} & \textbf{\#example} \\
\toprule
Task 1          & 4415    & 172495    & 626540 &       378720             \\
Task 2     & 425 & - & 913  &      18238          \\
\bottomrule
\end{tabular}
\caption{Supporting dataset.}
\label{tab:supporting_dataset_analysis}
\end{table}

For the finetuned task, we use a similar approach with our previous systems \cite{jnlp_task4_coliee2019, nguyen2020jnlp}. We use the Japanese Civil Code and the data given by COLIEE's organizer as training and validation data. Working on multilingual data, we create negation rules for Japanese and removed law sentences which are represented as a list. After augmentation, we obtain 7000 sentences in two languages.

\section{Experiments}
\subsection{Task 1 and Task 2. Case Law Processing}
\textbf{Training data} As we mentioned in Section~\ref{sec:method}, we create a supporting training dataset to train BERT model and the analysis of this training dataset is as Table \ref{tab:supporting_dataset_analysis}. From 4415 cases in Task~1 raw dataset, we can extract more than 170K paragraphs. However, inside these paragraphs, some of them contain a lot of french text. This actively demonstrates that we need a filter step to extract clean English text. For the purpose of avoiding noise in the dataset, we use langdetect\footnote{https://pypi.org/project/langdetect/} to filter and remove french text from training data. The clean paragraphs are obtained so far, we split every single paragraph into sentences (>625K in total), and from these sentences we apply some technique to generate supporting examples. The massive number of silver training examples for training BERT is over 350K examples.

Besides silver data, we utilize the data from Task~2 to extract more training dataset. As you can see in the Table \ref{tab:supporting_dataset_analysis}, the number of gold examples we can extract is more than 18K.

For Task~1, we submit 3 runs as follow:
\begin{itemize}
    \item Run1: Lexical score combine with semantic score with ratio $\alpha$ 7:3.
    \item Run2: Lexical score combine with semantic score with ratio $\alpha$ 3:7.
    \item Run3: Only supporting score.
\end{itemize}

For Task~2, we submit 3 runs as follow:

\begin{itemize}
    \item Run1: Lexical score combine with semantic score with ratio $\alpha$ 7:3.
    \item Run2: Lexical score combine with semantic score with ratio $\alpha$ 7:3 and finetuning on gold training dataset.
    \item Run3: Lexical score combine with NFSP model's score.
\end{itemize}

Our method on Task~1 has bad performance on the test set, the particular reason for this issue is the problem when we limit the searching space from 4000 to 100, maybe lexical matching works badly in this circumstance. For Task~2, as the numbers in Table \ref{tab:Task2_result}, our method achieves 61\% on F score, and we place 5\uppercase{th} in COLIEE 2021.

\begin{table}[]
\begin{tabular}{|l|r|}
\hline
 \textbf{Run ID}                     & \textbf{F1 Score}     \\ \hline
BM25Supporting\_Denoising            & 0.6116               \\ \hline
 BM25Supporting\_Denoising\_Finetune   & 0.6091                \\ \hline
 NFSP\_BM25                           & 0.5868               \\ \hline

\end{tabular}
\caption{Result on Task~2.}
\label{tab:Task2_result}
\end{table}

\subsection{Task 3. The Statute Law Retrieval Task}
We trained and evaluated our proposed models for this task with the previous year's dataset. Macro- precisions, recalls, and $F_{2}$ scores are reported. Note that the reported $F_{2}$ scores are the $F_{2}$ scores of class-wise precision means and class-wise recall means.

We conducted multiple experiments with different settings to find the most appropriate settings when applying text chunking technique. In these experiments, we finetuned 3 epochs on \textit{bert-base-japanese}\footnote{https://huggingface.co/cl-tohoku/bert-base-japanese} pretrained model and report results in Table~\ref{tab:task3_results_chunk}. We use <$window\_size$>/<$stride$> to denote the sliding window parameters. The results indicate that <$150$>/<$50$> seems to be the most appropriate setting for the task. We use this setting to conduct experiments relating to self-training technique. Specifically, first, we finetuned \textit{bert-base-japanese} pretrained model with $e_{1}$ epochs, then performed self-labeling process and continued to finetune with $e_{2}$ epochs. We use <$e_{1}$>/<$e_{2}$> to denote this settings. The results in Table~\ref{tab:task3_results_self_labeled} suggests that $e_{1} = 2$ seems to be the “just right” parameter for the first finetuning process (when $e_{1} = 3$, the finetuned model seems to start overfitting); and as $e_{2}$ increases, $F_{2}$ tends to increase. The results demonstrate the positive impact of our proposed methods: $F2$ increased from 71.62 to 72.66 when applying the text chunking technique, and this number is 72.91 with the self-labeled technique.

\begin{table}[]
\small
\begin{tabular}{lccccc}
\hline
\textbf{Chunking info.}&\textbf{Return}&\textbf{Retrieved}&\textbf{P}&\textbf{R}&\textbf{F2} \\
\hline
no chunking       &  118   &    81     &       68.24       &     72.52      &   71.62   \\
110/20 &  190   &  {\ul108}  &       61.20       &     67.87      &   66.42   \\
150/10 &  122   &    85     &       64.77       &     66.67      &   66.28   \\
150/20 &  129   &    93     &       68.09       &     70.72      &   70.18   \\
150/40 &  131   &    92     &       67.12       &     71.17      &   70.32   \\
150/50 &  132   &    94   &     {\ul69.74}     &   {\ul73.42}    & {\ul72.66}\\
200/50 &  139   &    90     &       67.12       &     72.97      &   71.72   \\
300/50 &  110   &    74     &       65.39       &     67.57      &   67.12   \\
\hline
\end{tabular}
\caption{(Task 3) Results of \textit{bert-base-japanese} pretrained model finetuning with the text chunking technique. The  values in the \textit{Chunking info.} column denotes chunking settings with format <$window\_size$>/<$stride$>.}
\label{tab:task3_results_chunk}
\end{table}

\begin{table}[]

\begin{tabular}{lccccc}
\hline
\textbf{Setting}&\textbf{Return}&\textbf{Retrieved}&\textbf{P}&\textbf{R}&\textbf{F2} \\
\hline
3/0&132&94&{\ul69.74}&73.42&72.66\\
1/1&{\ul109}&81&68.02&67.57&67.66\\
1/2&145&98&66.89&72.52&71.32\\
1/3&133&96&68.77&72.97&72.09\\
2/1&161&95&64.55&72.52&70.77\\
2/2&161&95&64.55&72.52&70.77\\
2/3&195&{\ul104}&62.39&{\ul76.13}&{\ul72.91}\\
3/1&146&97&63.32&68.92&67.72\\
3/2&146&96&64.37&71.17&69.70\\
3/3&147&97&60.02&65.77&64.53\\
\hline
\end{tabular}
\caption{(Task 3) Results of \textit{bert-base-japanese} pretrained model with the self-labeled technique.  The  values in the \textit{Setting} column follows the format <$e_{1}$>/<$e_{2}$>.}
\label{tab:task3_results_self_labeled}
\end{table}

\begin{table}[]
\small
\begin{tabular}{lccccc}
\hline
\textbf{Run ID}&\textbf{sid}&\textbf{F2}&\textbf{Prec}&\textbf{Recall}\\
\hline
OvGU\_run1&E/J&0.7302&0.6749&0.7778\\
{\ul JNLP.CrossLMultiLThreshold}&E/J&0.7227&0.6000&0.8025\\
BM25.UA&E/J&0.7092&0.7531&0.7037\\
{\ul JNLP.CrossLBertJP}&E/J&0.7090&0.6241&0.7716\\
R3.LLNTU&E/J&0.7047&0.6656&0.7438\\
R2.LLNTU&E/J&0.7039&0.6770&0.7315\\
R1.LLNTU&E/J&0.6875&0.6368&0.7315\\
{\ul JNLP.CrossLBertJPC15030C15050}&E/J&0.6838&0.5535&0.7778\\
OvGU\_run2&E/J&0.6717&0.4857&0.8025\\
TFIDF.UA&E/J&0.6571&0.6790&0.6543\\
LM.UA&E/J&0.5460&0.5679&0.5432\\
TR\_HB&E/J&0.5226&0.3333&0.6173\\
HUKB-3&J&0.5224&0.2901&0.6975\\
HUKB-1&J&0.4732&0.2397&0.6543\\
TR\_AV1&E/J&0.3599&0.2622&0.5123\\
TR\_AV2&E/J&0.3369&0.1490&0.5556\\
HUKB-2&J&0.3258&0.3272&0.3272\\
OvGU\_run3&E/J&0.3016&0.1570&0.7006\\
\hline
\end{tabular}
\caption{(Task 3) Result of final runs on the test set, the underlined lines refer to our models.}
\label{tab:task3_runs_results}
\end{table}

We also did experiments with \textit{bert-base-japanese-whole-word-masking}\footnote{https://huggingface.co/cl-tohoku/bert-base-japanese-whole-word-masking} and \textit{xlm-roberta-base}\footnote{https://huggingface.co/xlm-roberta-base} pretrained models, but we do not report in this paper because of paper length limitation. We performed model ensembling on the outputs of those models.

For the contest submissions, we submitted 3 runs based on the 3 proposed approaches. The model ensembling method returned the highest result among the three. We are the runner-up of this task. The results of final runs of all participants are in Table~\ref{tab:task3_runs_results}.

\subsection{Task 4. The Legal Textual Entailment Task}
Similar to Task 3,  we also trained and evaluated our proposed models with the dataset in the previous year with the questions having id \textit{R-01-*} as the development set. Because of the relatively small training data, we run 5 times with each setting and report the mean and standard deviation values. For this task, most of the hyperparameters follows settings in \citet{nguyen2020jnlp} where $batch\_size=16$ and $learning\_rate=1e^{-5}$.

\paragraph{\textbf{Pretrained model and Data augmentation}} Firstly, we conducted the experiments to find the most suitable pretrained model for this task, and the most suitable setting for the tf-idf-based data augmentation method  (Table~\ref{tab:t4_multi_model}). Based on the experimental results, we found that the \textit{bert-base-japanese-whole-word-masking} pretrained model is more suitable for this task than others. Besides, the tf-idf-based augmentation data method also help increase the model performance.

\begin{table}[]
\small
\begin{tabular}{lccccc }
\hline
\textbf{Model}&\textbf{Origin}& \textbf{tf-idf1} & \textbf{tf-idf2} & \textbf{tf-idf5} & \textbf{tf-idf20} \\
\hline
BertJp       &  $55.9\pm3.7$   &   -   &      -   &     -       &  -   \\
BertJp2       &  $60.4\pm4.1$   &   $61.6\pm5.0$      &        \textbf{$62.7\pm5.7$}     &     $61.3\pm4.4$       &   $58.4\pm2.9$    \\
\hline
\end{tabular}
\caption{(Task 4) Model accuracies with different tf-idf augmentation setting. The name \textit{BertJp}, \textit{BertJp2} indicates that we used a pre-trained \textit{bert-base-japanese} and \textit{bert-base-japanese-whole-word-masking}, respectively. The name column follows the format \textit{tf-idf<number>} where \textit{<number>} denotes the number of augmented articles appended for each question.}
\label{tab:t4_multi_model}
\end{table}

\paragraph{\textbf{Performance stability}} In addition, we found that the performance of the model with a small epoch is fairly unstable. Therefore, we experimented with a bigger number of training epochs (Table~\ref{tab:t4_epoch}). The experimental results demonstrate that the runs with higher epochs tend to achieve more stable accuracies, but the model can be overfitted if we increase the number of epochs too much. Besides, the augmentation data also helps the model performance to be more stable.

\begin{table}[]
\small
\begin{tabular}{cccc }
\hline
\textbf{\# epochs}&  \textbf{tf-idf1} & \textbf{tf-idf2} & \textbf{tf-idf5}  \\
\hline
3       &   $61.6\pm5.0$      &         $62.7\pm5.7$     &     $61.3\pm4.4$      \\
10       &  $61.8\pm3.0$   &              $62.9\pm2.2$      &        \uline{$64.7\pm2.5$}    \\
20      &   -  &   $61.6\pm1.8$      &         $64.0\pm1.3$       \\
\hline
\end{tabular}
\caption{(Task 4) Accuracies of \textit{BertJp2} with different training epochs.}
\label{tab:t4_epoch}
\end{table}

\paragraph{\textbf{Long article challenge.}} Finally, to address the long article challenge, we conducted the experiments using the \textit{text chunking} and the \textit{self-labeled} techniques described in Task 3 (Table~\ref{tab:t4_self_lb}). Although the accuracy of models in this setting did not increase, the variant ranges are smaller. It may be because the \textit{text chunking} and the \textit{self-labeled} techniques help eliminate noises in training data.

\begin{table}[]
\begin{tabular}{lcc }
\hline
\textbf{Setting}&   \textbf{tf-idf2} & \textbf{tf-idf5}  \\
\hline
1/10       &           $62.9\pm1.7$     &     $63.1\pm1.9$      \\
2/10       &           \uline{$64.3\pm0.5$ }    &        \uline{$64.3\pm0.5$ }    \\
3/10      &           $62.2\pm1.7$      &         $64.1\pm1.0$       \\
\hline
\end{tabular}
\caption{(Task 4) Accuracies of \textit{BertJp2} using the \textit{self-labeled} technique with different settings on chunking data where \textit{window\_size = 150}, \textit{stride = 50}.  The  values in the \textit{Setting} column denotes <$e_{1}$>/<$e_{2}$>.}
\label{tab:t4_self_lb}
\end{table}

The results on blind test set are shown in the Table~\ref{tab:t4_test_rc} with the id ``$JNLP.Enss5C15050$'' refers to the model BertJp2 using augmentation data tf-idf5; ``$JNLP.Enss5C15050SilverE2E10$'' refers to the model BertJp2 using augmentation data tf-idf5, and <$e_{1}$>/<$e_{2}$> is 2 /10; ``$JNLP.EnssBest$'' refers to the enssemble of both models. 

\begin{table}[]
\begin{tabular}{|l|l|c|c| }
\hline
\textbf{Team}&   \textbf{sid} & \textbf{Correct} & \textbf{Acc.}  \\
\hline
HUKB    &      HUKB-2       &    57 & 0.7037      \\
UA       &         UA\_parser  &       54  & 0.6667\\
{\ul JNLP}      &           JNLP.Enss5C15050  &      51 & 0.6296    \\
{\ul JNLP}     &           JNLP.Enss5C15050SilverE2E10   &        51 & 0.6296      \\
{\ul JNLP}      &          JNLP.EnssBest  &     51 & 0.6296   \\
OVGU &          OVGU\_run3      &      48 & 0.5926     \\
TR &          TR-Ensemble      &      48 & 0.5926     \\
KIS &           KIS1      &        44 & 0.5432        \\
UA &           UA\_1st     &        44 & 0.5432        \\
\hline
\end{tabular}
\caption{(Task 4) Results final runs  on the test set. The underlined lines refer to our submissions.}
\label{tab:t4_test_rc}
\end{table}

\subsection{Task 5. Statute Law Question Answering}
We compare our proposed models together and with other cross-lingual and multilingual baselines such as XLM-RoBERTa \cite{conneau2019unsupervised} and original BERT Multilingual~\cite{devlin2018bert}. In the 7000 augmented sentences, we divide the train set and validation set with the ratio of 9:1.

Our experiments show that NFSP Base and NMSP Base achieve the best performance and have stable loss decrease. NFSP Distilled, NMSP Distilled and XLM-RoBERTa fail to learn from the data and their performance equivalent to that of random sampling. BERT Multilingual is in the middle of the ranked list of models in Table~\ref{tab:finetune_result_validation}. Therefore, we choose NFSP Base, NMSP Base and original BERT Multilingual as candidates for the final run.

\begin{table}[]
\begin{tabular}{lr}

\toprule
\textbf{Model}    & \textbf{Accuracy} \\
\toprule
NFSP Base          & 71.0\%                       \\
NFSP Distilled     & 51.1\%                       \\
\midrule
NMSP Base          & 79.5\%                       \\
NMSP Distilled     & 48.9\%                       \\
\midrule
XLM-RoBERTa       & 51.1\%                       \\
BERT Multilingual & 64.1\%                      \\
\bottomrule
\end{tabular}
\caption{(Task 5) Performance of models on validation set.}
\label{tab:finetune_result_validation}
\end{table}

Table~\ref{tab:final_runs} is the result of the models on the blind test of COLIEE-2021's organizer. On this test set, NFSP Base outperforms other methods and becomes the best system for this task. NMSP is in third place and the original BERT Multilingual has the performance below the baseline. These results again support our proposal in pretraining models using sentence-level cross-lingual information.

\begin{table}[]
\small
\begin{tabular}{|l|l|l|r|}
\hline
\textbf{Team} & \textbf{Run ID}                     & \textbf{Correct} & \textbf{Accuracy}     \\ \hline
              & BaseLine                            & No 43/All 81     & 0.5309                \\ \hline
{\ul JNLP}    & {\ul JNLP.NFSP}                     & {\ul 49}         & {\ul \textbf{0.6049}} \\ \hline
UA            & UA\_parser                          & 46               & 0.5679                \\ \hline
{\ul JNLP}    & {\ul JNLP.NMSP}                     & {\ul 45}         & {\ul 0.5556}          \\ \hline
UA            & UA\_dl                              & 45               & 0.5556                \\ \hline
TR            & TRDistillRoberta                    & 44               & 0.5432                \\ \hline
KIS           & KIS\_2                              & 41               & 0.5062                \\ \hline
KIS           & KIS\_3                              & 41               & 0.5062                \\ \hline
UA            & UA\_elmo                            & 40               & 0.4938                \\ \hline
{\ul JNLP}    & {\ul JNLP.BERT\_Multilingual} & {\ul 38}         & {\ul 0.4691}          \\ \hline
KIS           & KIS\_1                              & 35               & 0.4321                \\ \hline
TR            & TRGPT3Ada                           & 35               & 0.4321                \\ \hline
TR            & TRGPT3Davinci                       & 35               & 0.4321                \\ \hline
\end{tabular}

\caption{(Task 5) Result of final runs on the test set, the underlined lines refer to our models.}
\label{tab:final_runs}
\end{table}

\section{Conclusions}
This paper presents JNLP Team's approaches using deep learning to the legal text processing tasks in the COLIEE-2021 competition.
Due to the limited amount of data and the difficulty of the tasks, we used pretraining methods to solve the problems.
The experimental results and the performance of the models on the blind test show the reasonability and robustness of the proposed methods.

\begin{acks}
This work was supported by JSPS KAKENHI Grant Numbers JP17H06103 and
JP20K20406.
\end{acks}

\bibliographystyle{ACM-Reference-Format}
\bibliography{references}

\appendix

\end{document}